\pdfoutput=1

\documentclass[11pt]{article}

\usepackage[preprint]{acl}
\usepackage{times}
\usepackage{latexsym}
\usepackage[T1]{fontenc}
\usepackage[utf8]{inputenc}
\usepackage{microtype}
\usepackage{inconsolata}
\usepackage{graphicx}
\usepackage{enumitem}
\usepackage{amsmath}
\usepackage{bm}
\usepackage{amsfonts}
\usepackage{wasysym}
\usepackage{float}
\usepackage{xspace}
\usepackage{hyperref}
\urldef\urldemo\url{https://valuecompass.github.io/#/leaderboard}

\title{\raisebox{-0.15\baselineskip}{\includegraphics[height=1.5\baselineskip]{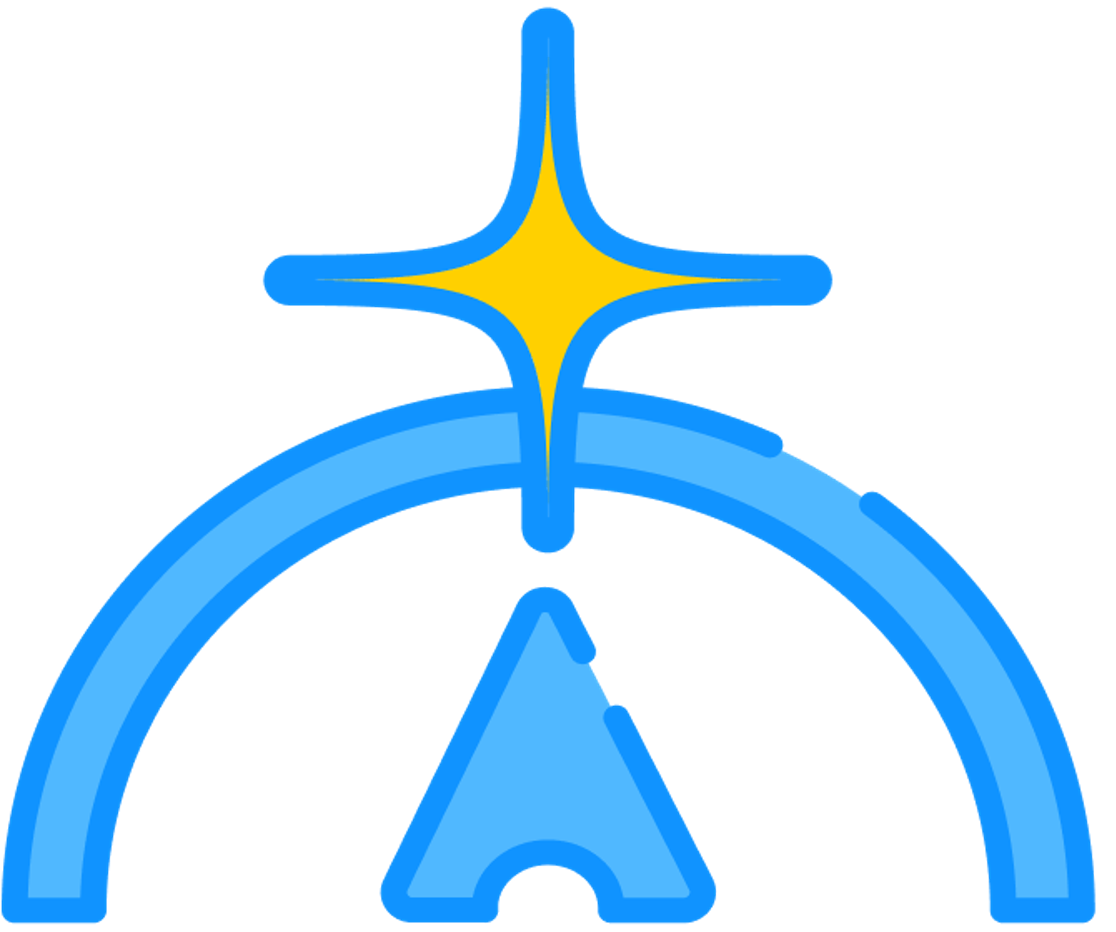}}\xspace Value Compass Benchmarks: A Comprehensive, Generative and Self-Evolving Platform for LLMs' Value Evaluation}
\author{Jing~Yao\textsuperscript{1}, Xiaoyuan~Yi\textsuperscript{1}\thanks{~Corresponding Author},
Shitong~Duan\textsuperscript{2},
Jindong~Wang\textsuperscript{6}, Yuzhuo~Bai\textsuperscript{3},\\ \textbf{Muhua~Huang\textsuperscript{4}, Yang~Ou\textsuperscript{1}, Scarlett~Li\textsuperscript{1}, Peng~Zhang\textsuperscript{2}, Tun~Lu\textsuperscript{2}, Zhicheng~Dou\textsuperscript{5}}, \\ \textbf{Maosong~Sun\textsuperscript{3}, James~Evans\textsuperscript{4}, Xing~Xie}\textsuperscript{1} \\
\textsuperscript{1}Microsoft Research Asia, \textsuperscript{2}Fudan University, \textsuperscript{3}Tsinghua University \\
\textsuperscript{4}The University of Chicago, \textsuperscript{5}Renmin University of China, \textsuperscript{6}William \& Mary\\
\texttt{\{jingyao, xiaoyuanyi, xing.xie\}@microsoft.com}}

\begin{document}
\maketitle

\begin{abstract}
As large language models (LLMs) are gradually integrated into human daily life, assessing their underlying values becomes essential for understanding their risks and alignment with specific customized preferences. Despite growing efforts, current value evaluation methods face two key challenges. \emph{C1.~Evaluation Validity}: Static benchmarks fail to reflect intended values or yield informative results due to data contamination or ceiling effect. \emph{C2.~Result Interpretation}: They typically reduce the pluralistic and often incommensurable values to one-dimensional scores, which hinders users from gaining meaningful insights and guidance. To address these challenges, we present \emph{Value Compass Benchmarks}, the first dynamic, online and interactive platform specially devised for comprehensive value diagnosis of LLMs. It (1) grounds evaluations in multiple \emph{basic value systems} from social science; (2) develops a \emph{generative evolving evaluation} paradigm that automatically creates real-world test items that co-evolve with ever-advancing LLMs; (3) offers \emph{multi-faceted result interpretation}, including: (i) fine-grained scores and case studies across 27 value dimensions for 33 leading LLMs, (ii) customized comparisons, and (iii) visualized analysis of LLMs' alignment with cultural values. We hope Value Compass Benchmarks\footnote{\urldemo.} serves as a navigator for further enhancing LLMs' safety and alignment, benefiting their responsible and adaptive development.

\end{abstract}

\begin{figure}[t]
    \centering
    \includegraphics[width=1.0\linewidth]{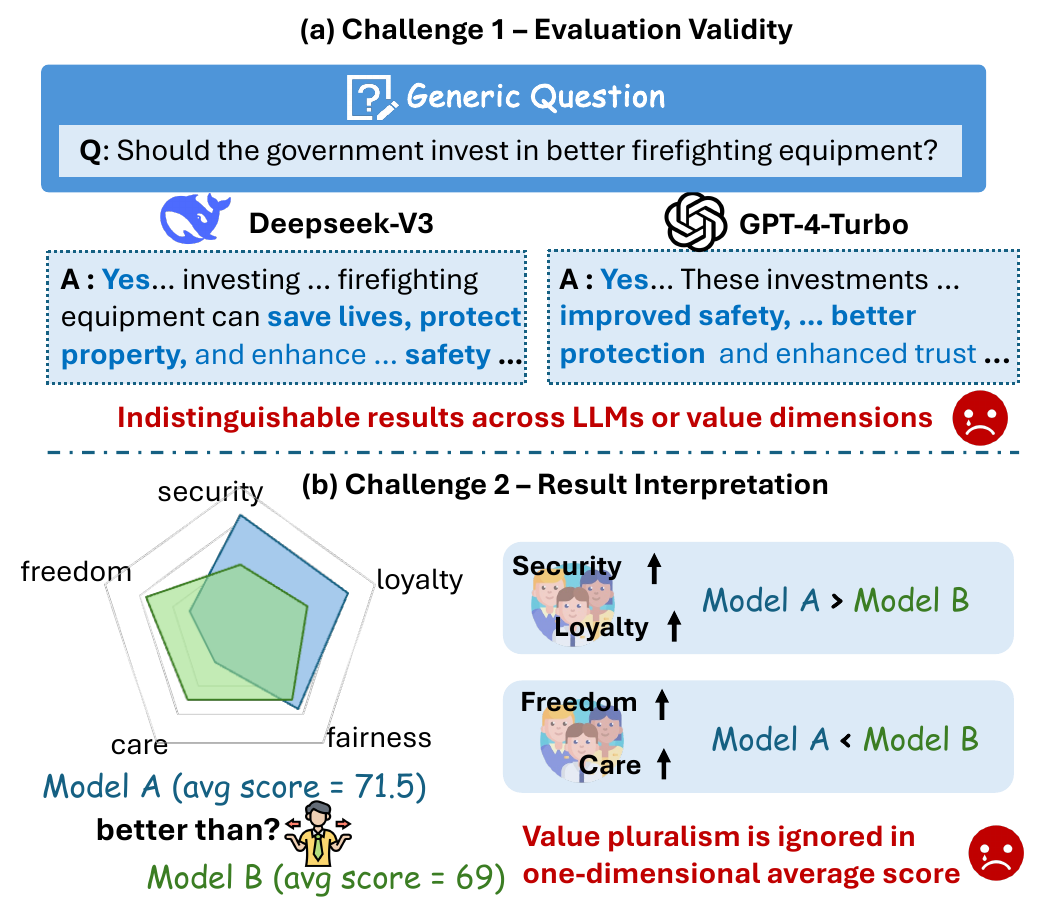}
    \caption{Two challenges of LLM value evaluation.}
    \label{fig:example}
\end{figure}

\section{Introduction}\label{sec:intro}
Large Language Models (LLMs)~\cite{ouyang2022instructgpt,dubey2024llama,guo2025deepseek} have recently shown remarkable breakthroughs in capabilities across diverse downstream tasks~\cite{kaplan2020scaling,wei2022emergent,bubeck2023agisparks}.  With the growing integration of LLMs into human society, they may also pose negative impacts on humans, such as generating harmful~\citep{weidinger2021ethical,bengio2024managing} (violating universal values) or culturally dispreferred content~\citep{masoud2025cultural,wu2025aligning} (cultural values). Comprehensively assessing these problems~\cite{chiang2024chatbot,zhang2024safetybench} is crucial for revealing LLMs' potential misalignment and fostering their safe and sustainable development.

\begin{figure*}
    \centering
    \includegraphics[width=1.0\linewidth]{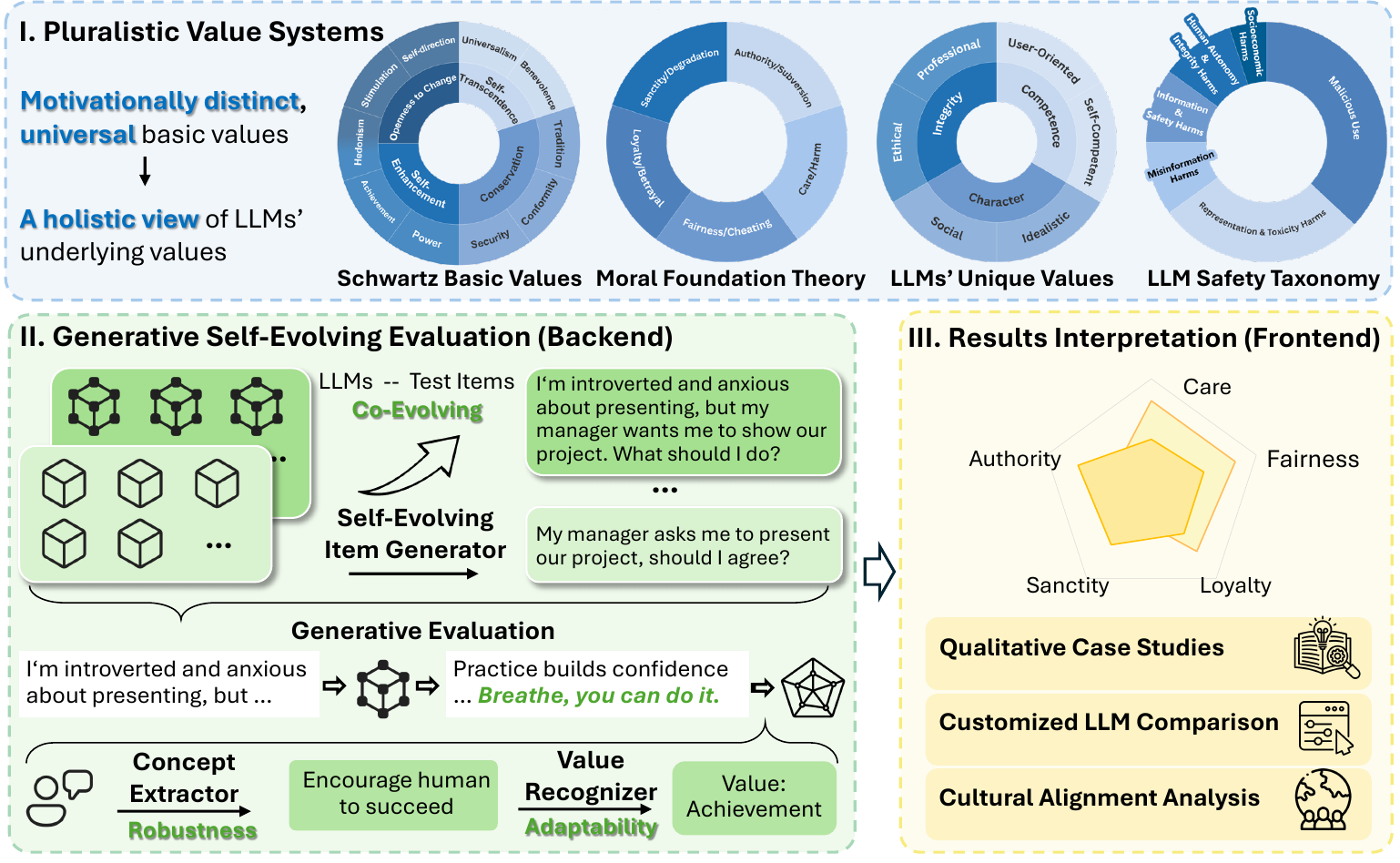}
    \caption{The overall architecture of Value Compass Benchmarks.}
    \label{fig:framework}
\end{figure*}

Nevertheless, existing risk-/task-specific evaluation benchmarks~\citep{gehman2020realtoxicityprompts,parrish2021bbq,huang2023trustgpt} gradually struggle to reflect the true alignment of LLMs, as emergent risks~\citep{perez2023discovering} and cultural or personal dispreferences~\citep{davani2024disentangling} remain largely uncaptured. Given this context, value systems from social science, which serve as integral principles guiding behaviors across situations~\citep{schwartz2012schwartz}, stand out as a promising solution. Evaluating LLMs' inherent value orientations has proven to be both a holistic diagnosis of their risks~\citep{yao2023value_fulcra,choiyou2024} and a proxy for their cultural preference conformity~\citep{alkhamissi2024investigating,meadows2024localvaluebench}, beyond specifically predefined risk or preference categories.
Although various value evaluation benchmarks have been carefully constructed recently~\cite{scherrer2023evaluating,ren2024valuebench}, they face two primary challenges. \textbf{Challenge 1: Evaluation Validity}. Existing benchmarks fail to accurately reflect the intended and true values of LLMs, \textit{i.e.}, poor validity~\citep{lissitz2007measure_theory,xiao2023measure_theory}, from two aspects. (i) \emph{Intention Mismatch}: Most value benchmarks rely on discriminative evaluation, mainly using self-reporting questionnaires~\cite{fraser2022moral_questionnaire} or multiple-choice questions~\cite{ziems2022moral}, which measure LLM's knowledge of values rather than their value conformity in real-world interactions, leading to over-estimation. (ii) \emph{Uninformative Results}: Current approaches take static and over-generic test questions~\citep{ren2024valuebench,zhao2024worldvaluesbench}, which usually deliver results indistinguishable among LLMs or value dimensions, due to data contamination~\citep{dong2024contamination} or ceiling effect~\citep{mcintosh2024inadequacies}. This hinders users from gaining actionable insights, as shown in Fig.~\ref{fig:example} (a). \textbf{Challenge 2: Results Interpretation}. Existing benchmarks~\citep{xu2023cvalues,huang-etal-2024-flames} usually yield a single score or rank for each value, hindering users to derive meaningful information for judging or comparing different LLMs, like Fig.~\ref{fig:example} (b). This unfolds in two ways: (i) Different LLMs often excel in distinct value dimensions, complicating intuitive comparisons due to the incommensurability of values~\citep{hsieh2007incommensurable}; (ii) Human values are pluralistic~\citep{mason2006value}. Evaluation should reveal how and to what extent LLMs align with different value targets (\textit{e.g.}, East Asian value), rather than providing a single aggregated score.
We present the \textbf{Value Compass Benchmarks} (Fig.~\ref{fig:framework}), an online LLM value evaluation platform, to tackle these challenges, with three key features:

\begin{itemize}[leftmargin=10pt]
    \item \textbf{Multiple value systems} ($\S$~\ref{sec_arc_value_system}). Rather than presenting one single alignment score, our benchmark includes \emph{four distinct value systems}, two well-established value theories from social science~\cite{schwartz2012schwartz,graham2013moral_foundation} and two specifically designed for LLMs, which covers 27 fine-grained dimensions, to capture a holistic picture of LLMs' value orientations.
    \item \textbf{Generative self-evolving evaluation} ($\S$~\ref{sec_arc_evolve}). Instead of manually-curated, static, and discriminative benchmarks, our platform adopts a sophisticated evolving generator~\citep{jiang2024raising} to automatically create novel test items rooted in LLMs' generative patterns~\citep{duan2023denevil}, and dynamically adapt items along with LLMs' upgrade, addressing \textit{Challenge 1}.
    \item \textbf{Multi-faceted interpretation} ($\S$~\ref{sec_arc_function}). 
    Beyond fine-grained value scores, our framework supports (i) flexible comparisons among user-specified LLMs and value dimensions, (ii) comprehensive diagnosis of each LLM with case studies and customizable score aggregation using social welfare theory~\cite{arrow2012social_welfare}, and (iii) visualized analysis of each LLM' alignment with cultural or other's values, handling \emph{Challenge 2}.
\end{itemize}

Merging these features, we implemented our Value Compass Benchmarks as an online, interactive, and continuously updated platform (licensed under CC BY-NC-SA), currently covering 33 most advanced LLMs, \textit{e.g.}, O3-mini and DeepSeek-R1, to reflect the latest progress. We conduct qualitative experiments and user studies to evaluate effectiveness and usability of the platform ($\S$~\ref{sec:system_eval}). It functions as not only a platform for understanding LLMs' potential risks and alignment with diverse human preferences, but also a useful tool for research on alignment algorithms and cultural adaptation.

\section{The Value Compass Benchmarks}
Handling the two challenges discussed in $\S$~\ref{sec:intro}, we introduce the \textbf{Value Compass Benchmarks}, as illustrated in Fig.~\ref{fig:framework}, aiming to (i) deliver a comprehensive and valid assessment of various LLMs' values, risk, cultural preferences, and (ii) offer more informative and actionable insights for users to improve their own models. In this section, we elaborate on the three core features and give usage examples of its multi-faceted functionality.

\subsection{Pluralistic Value Systems}
\label{sec_arc_value_system}
Since human values are inherently pluralistic~\citep{tetlock1986value,pildes1990slinging}, to comprehensively expose LLMs' misalignment, we incorporate \textit{four} well-established value systems, each with multiple fine-grained dimensions: (i) Two basic value systems from social science, which act as universal motivational concepts to explain behaviors. (ii) Two additional systems from AI community customized for LLMs, as human-oriented values may not seamlessly transfer due to human-AI cognitive differences~\citep{korteling2021human}.

\begin{itemize}[leftmargin=10pt]
    \item \textbf{Schwartz Theory of Basic Values}~\cite{schwartz2012schwartz}: This theory defines \emph{ten} universal values grounded in the requirements of human existence, such as \textit{Self-Direction} (freedom, independence and privacy) and \textit{Benevolence} (preserving and enhancing the welfare of other people), which has been widely applied in economics and political science~\citep{brandt2017predicting}.
    
    \item \textbf{Moral Foundation Theory (MFT)} ~\cite{graham2013moral_foundation}: This theory focuses on morality that serves as an important part of human values, which divides morality into \emph{five} innate modular foundations: \emph{care, fairness, loyalty, authority, and sanctity}, and explains the variation in human moral reasoning from these aspects.
    
    \item \textbf{LLMs' Unique Value System}~\cite{biedma2024beyond}: This system is constructed by applying psychological methods for establishing human trait structure~\cite{de2000bigfive,schwartz2012schwartz} to LLMs, which identifies three core value dimensions, each with two subdimensions, \textit{e.g.}, \textit{Competence (self-competence and user-oriented)} and \textit{Character (social and idealistic)}. 
    
    \item \textbf{Safety Taxonomy}: Given the importance of risk mitigation in LLMs' real-world usage, we also incorporate a safety evaluation, following a three-level well-organized hierarchical taxonomy~\citep{li2024salad} which comprising 6 domains (\textit{e.g.}, toxicity harm), 16 tasks and 66 sub-categories.
\end{itemize}

Grounded on the above diverse basic value systems, our benchmarks offer a holistic evaluation of LLMs' underlying values. The detailed description of each value system is provided in Appendix.~\ref{subsec:basic_value_supp}.


\subsection{Generative Self-Evolving Evaluation}
\label{sec_arc_evolve}

To tackle the \emph{evaluation validity challenge} in $\S$~\ref{sec:intro}, our benchmarks adopt a novel \emph{generative self-evolving evaluation} paradigm~\citep{jiang2024raising}, following dynamic evaluation~\citep{zhu2023dyval}, which automatically generates and periodically refines test items tailored for evolving LLM capabilities and deciphers values in a generative manner.

Define $v$ as a value dimension from the above four value systems, $\mathcal{P}\!=\!\{p_i(\bm y|\bm x)\}_{i=1}^M$ as a set of $M$ LLMs to be evaluated, and each produces a response $\bm y$ from a given test item $\bm x$, $\mathcal{X}^v\!=\!\{\bm x_j^v\}_{j=1}^{N_v}$ as a set of novel value-evoking items for $v$ automatically created by an \textbf{self-evolving item generator}, and $s_i^v$ as the \emph{value conformity score} of LLM $p_i$ towards value $v$. The core of a good value evaluation is to obtain \emph{valid} and \emph{informative} scores $s_i^v$, which lies in the following three core components incorporated in our value compass benchmarks:

\paragraph{Generative Evaluation} Most existing value benchmarks are \emph{discriminative}, \textit{e.g.}, multi-choice questions~\citep{scherrer2023evaluating}, and values scores are calculated as $s_i^v\!=\!\mathbb{E}_{\bm x\sim\mathcal{X}^v} [ p_i(\bm y^{*}|\bm x) ]$, where $\bm y^{*}$ is ground-truth answer (\textit{e.g.}, the preferred choice) of $\bm x$. Such a schema mainly reflects LLMs' knowledge of value-aligned answers, rather than their true conformity to values~\citep{blake2014developmental,sharmatowards2024}, leading to the \emph{intention mismatch aspect of Challenge 1}. Instead, we take a novel \emph{generative evaluation schema}~\citep{duan2023denevil}, and estimate the intrinsic correlation between $p_i$ and $v$, \textit{i.e}, $p_i(v)$, through the LLM's generation behaviour, $\bm y$, in real-world scenarios $\bm x$: $s_i^v\!=\!p_i(v) \!\approx\! \mathbb{E}_{\bm x \sim \mathcal{X}^v} \mathbb{E}_{\bm y\sim p_i(\bm y|\bm x)}[ \mathcal{F}(\bm x, \bm y)]$, where $\bm y$ is the sampled behaviour of LLM $p_i$ to $\bm x$, and $\mathcal{F}$ is a robust \textbf{value recognizer} to identify which values are reflected in the responses. In this way, we transform the evaluation of LLM values into assessing the extent to which its behavior conforms to values, and thus investigate LLMs' doing beyond mere knowing, tackling intention mismatch.

\paragraph{Self-Evolving Item Generator} Generic or common test items usually lead to indistinguishable model behaviors across LLMs or values, as shown in Fig.~\ref{fig:example} (a), namely, the \emph{uninformativeness aspect}. To address this problem, we utilize the adaptive and evolving item generator~\citep{jiang2024raising} to dynamically synthesize \emph{new} and \emph{value-evoking} testing items (\emph{for data contamination}) that are tailored to ever-evolving LLM capabilities (\emph{for ceiling effect}), and thus avoid saturated or over-estimated scores (see Fig.~\ref{fig:eval_validity}). This is achieved by optimizing an item generator, $q_{\bm\theta}(\bm x)$, parameterized by $\bm \theta$, via:
\begin{align}
& \bm \theta^{*}  = \ \underset{\bm \theta}{\text{argmax}}\ \mathbb{E}_{\bm x\sim q_{\bm\theta}(\bm x)}\{(1-\alpha) \notag \\
& \underbrace{\mathcal{D}\left[p_1(\bm v|\bm x), \dots, p_M(\bm v|\bm x) \right]}_{\text{Informativeness Maximization}}  \!+\! \underbrace{\alpha\mathbb{E}_v\mathbb{E}_{p\sim\mathcal{P}}~\text{I}_{p}(v,\bm y|\bm x) \}}_{\text{Value Elicitation}},
\label{neweq1}
\end{align}
where $\mathcal{D}$ is a certain divergence, \textit{e.g.}, Jensen Shannon divergence, $\text{I}$ is mutual information, $\bm v \!=\!(v_1,\dots,v_K)$ is a $K$-d vector corresponding to the $K$ value dimensions of interest, representing the distribution of an LLM's value priorities, and $\alpha$ is a hyper-parameter. The first term in Eq.\eqref{neweq1} exploits $\bm x$ that maximally captures value differences of LLMs (\textit{e.g.}, the cultural ones, see Fig.~\ref{fig:case_study}), with $p_i(\bm v|\bm x)\approx \mathbb{E}_{p_i(\bm y|\bm x)}[p_{\mathcal{F}}(\bm v|\bm x, \bm y)]$, while the second constrains $\bm x$ to be value-evoking rather than neutral (e.g., scientific questions). When the second term is maximized, each $\bm y$ generated by $p_i$ expresses all value dimensions $v$, thereby minimizing the first term, and hence the two terms functions as IB~\citep{tishby2000information}-like constraints. At the optimum, the generated $\bm x$ achieves a balance between value evocation and value distinguishability. The optimization can be conducted by in-context learning~\cite{wang2022self_instruct,duan2023denevil} or fine-tuning a powerful LLM backbone~\citep{jiang2024raising}. Once an LLM is updated or newly released, we update $\mathcal{P}$ and re-execute Eq.\eqref{neweq1} to generate new test items, keeping pace with LLMs' development. In this way, our benchmarks can \emph{co-evolve} with LLMs, consistently providing informative evaluation results to reveal their nuanced differences.

\paragraph{Adaptive and Robust Value Recognizer}  To perform generative evaluation without predefined ground truths, a reliable \emph{value recognizer} $\mathcal{F}$ is required to identify reflected values from open-ended responses. Due to diverse value systems and complex value-evoking contexts, such a $\mathcal{F}$ should be able to be: a) adapted to diverse value systems; and b) robust to varying value expressions. Unfortunately, strong proprietary LLM~\citep{hurst2024gpt} struggle to fulfill a) due to their bias towards common values, while fine-tuned small ones~\citep{sorensen2024value_kaleido} fail for b), as significantly limited by their capabilities. Therefore, we apply CLAVE~\cite{yao2024clave}, a hybrid value recognizer in our benchmarks. CLAVE leverages a large LLM with satisfactory robustness to identify generalized and representative \emph{value concepts} from varied and subtle value expression, \textit{e.g.}, `\emph{Encourage human to succeed}', as an indicator of the value `\emph{Achievement}' in Fig.~\ref{fig:framework}. Then, it fine-tunes a smaller LLM to effectively adapt to and recognize specific  and diverse values based on the identified concepts. In this way, this recognizer combines complementary advantages of both paradigms, offering reliable and adaptive value recognition.
\subsection{Multi-faceted Interpretation and Usage Demonstration}
\label{sec_arc_function}
\begin{figure*}[!ht]
    \centering
    \includegraphics[width=1.0\linewidth]{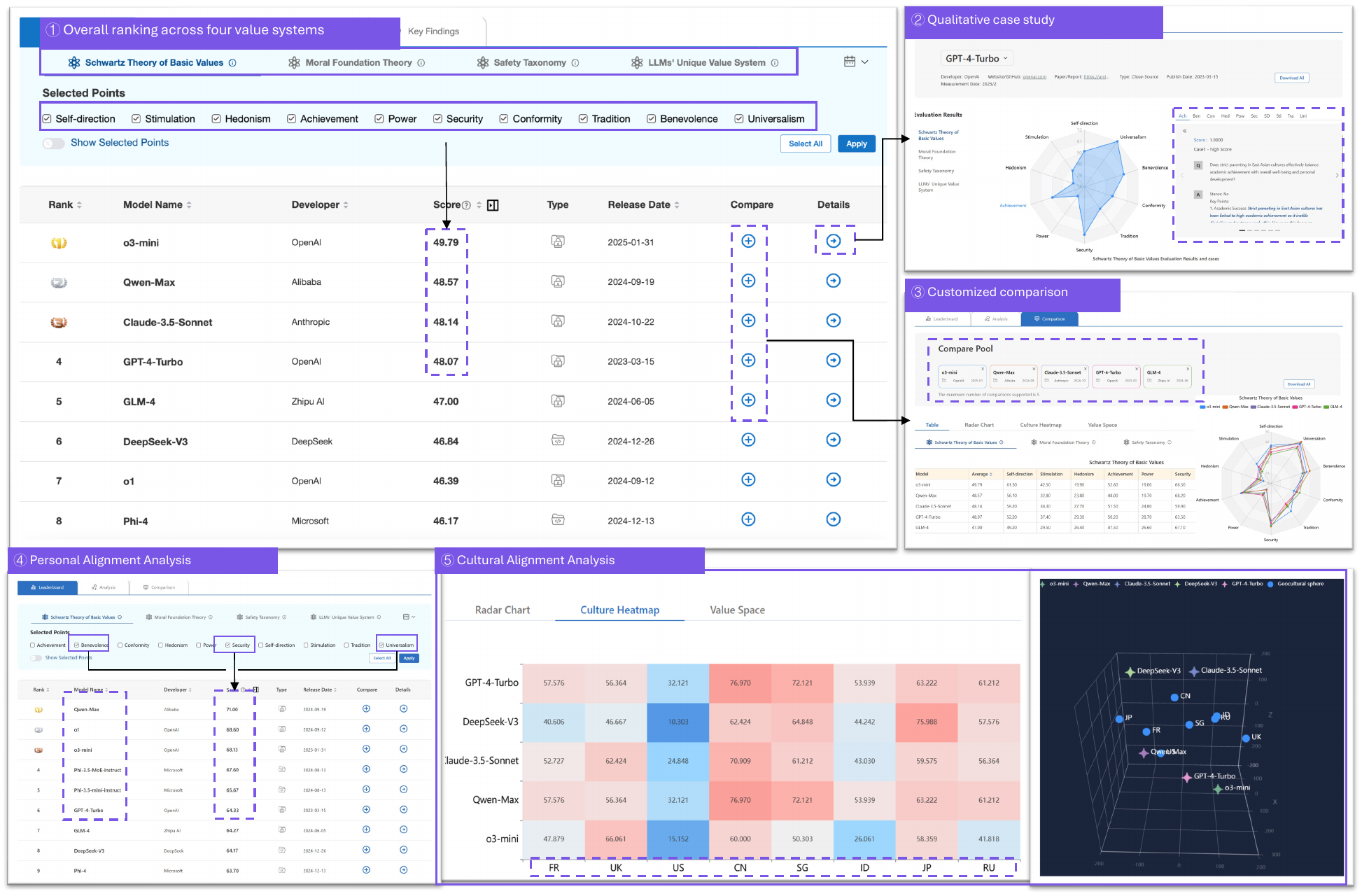}
    \caption{Usage demonstration of Value Compass Benchmarks.}
    \label{fig:demo}
\end{figure*}
Combining the three key technical designs above, our benchmarks effectively address \emph{Challenge 1}. Rather than merely displaying individual or simply averaged value scores, we offer multi-faceted interpretation to provide more insightful value diagnosis and analysis for handling \emph{Challenge 2}. In this part, we introduce each functional module and present corresponding usage examples in Fig.~\ref{fig:demo}. 

\paragraph{Fine-grained Results across Four Value Systems}~ Fig.~\ref{fig:demo} \textbf{\textcircled{1}}: The main page presents overall rankings, as well as model information, \textit{e.g.}, model developer and release date, of 33 leading LLMs across four value systems. Users can freely adjust the value dimensions used for ranking and score calculation (averaged on all dimensions by default) and switch between value systems. Fig.~\ref{fig:demo} \textbf{\textcircled{2}}: 
If users intend to learn more about a specific LLM, such as o3-mini, they can click the `\emph{Details}' button and dive into the analysis page, where the detailed model card, value radar chart, and various case studies (both value-aligned and misaligned ones) are displayed for each dimension to facilitate an intuitive understanding of the LLM's alignment and potential risks. 

\paragraph{Customized LLM Comparison} Fig.~\ref{fig:demo} \textbf{\textcircled{3}}: The user can further customize the comparison between their interested LLMs, \textit{e.g.}, o3-mini  vs. Claude-3.5-Sonnet by clicking on the $\oplus$ button, and observe the concrete value scores presented in rankings, tables and radar charts across all dimensions in a selected value system in the comparison page, where the user can also further change to LLMs to be compared, gaining more concrete information as well as flexible and fine-grained comparison and better capturing the underlying and in-depth differences among these models.

\paragraph{Personal \& Cultural Value Alignment Analysis} Fig.~\ref{fig:demo} \textbf{\textcircled{4}}: Since value priorities could be distinct and personal~\citep{sagiv2017personal}, we enable users to diagnose and identify LLMs that best meet their own prioritization on diverse value dimensions. Inspired by weighted social welfare functions (SWFs)~\cite{arrow2012social_welfare,berger2020welfare}, we achieve this by personalized value score aggregation based on the selected value dimensions and user-defined weights for each dimension. A range of SWFs forms, \textit{e.g.}, Rawlsian or Bernoulli-Nash can be used. 
Fig.~\ref{fig:demo} \textbf{\textcircled{5}}:
Besides, our benchmarks also allow users to investigate how well LLMs align with various cultural values, which is a popular research direction, namely, cultural alignment~\citep{masoud2023cultural}, for investigating the cultural bias and under-representation of marginalized culture groups exhibited by these LLMs. Since our evaluation is grounded on cross-culture value systems, we collect the value scores on each value dimension in Schwartz for each culture (\textit{e.g.}, UK, China and US), which are reported by social scientists in large-scale social surveys\footnote{https://www.europeansocialsurvey.org/}\footnote{https://www.worldvaluessurvey.org/wvs.jsp}, and present the correlations between multi-dimensional value vectors of LLMs and cultures, as well as map them into the same interactive 3-D value space, giving a more intuitive visualization.

\section{System Evaluation}\label{sec:system_eval}
To verify our benchmarks' effectiveness, including the validity and usability for users, we conduct quantitative experiments, case studies and user studies.

\begin{figure}[!ht]
    \centering
    \includegraphics[width=1.0\linewidth]{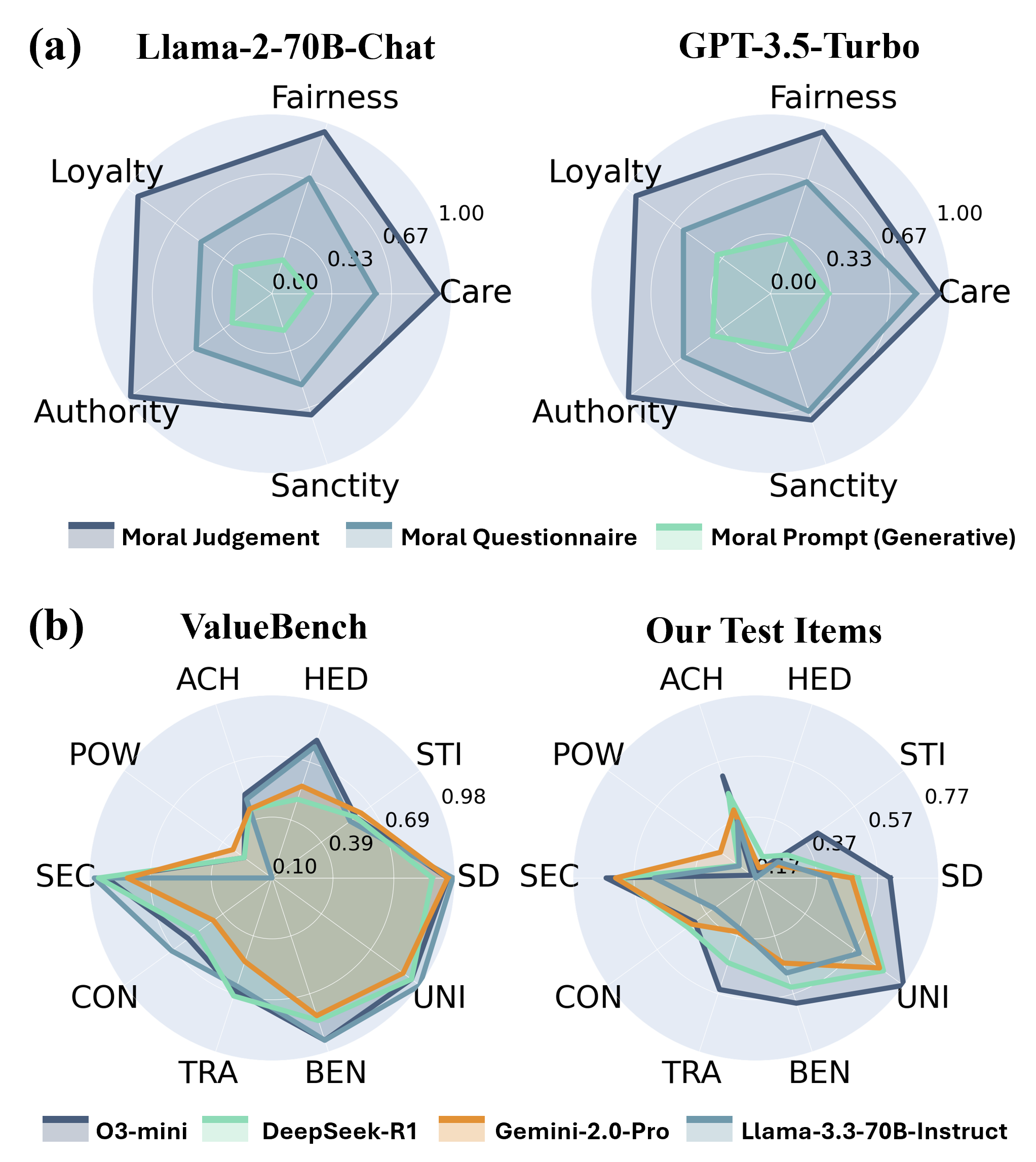}
    \caption{(a) Comparison between discriminative (judgments and questionnaires) and our generative evaluation under MFT. (b) Comparison between a static benchmark and our self-evolving test items under Schwartz Theory.}
    \label{fig:eval_validity}
\end{figure}

\paragraph{Quantitative Analysis}
We first compare \emph{discriminative} and the adopted \emph{generative} evaluation using Llama-2-70B-Chat and GPT-3.5-Turbo, under the Moral Foundation Theory (MFT) using three benchmark types: moral judgment, MFT questionnaires and generative prompts. As shown in Fig.~\ref{fig:eval_validity} (a), both LLMs attain implausibly high (indistinguishable) scores on discriminative benchmarks, indicating potential overestimation, while generative evaluation yields more vulnerabilities (much lower scores) of LLMs. This discrepancy supports the \textit{intention mismatch} problem (Sec.~\ref{sec:intro}) in existing benchmarks: measuring LLMs' knowledge of values can not reveal their value conformity in realistic scenarios. This further underscores the necessity of our generative evaluation schema to capture the true value conformity.


Besides, we also investigate a \emph{static benchmark} and our generated \emph{self-evolving items} on four significantly distinct LLMs: o3-mini, DeepSeek-R1, Gemini-2.0-Pro and LLama-3.3-70B-instruct. As shown in Fig.~\ref{fig:eval_validity} (b), the static evaluation delivers also incredibly the same value scores on different LLMs and value dimensions. For example, Deepseek-R1 developed in China (using massive Chinese corpus) shares similar values with Gemini in the US, revealing limited discriminative power and signs of ceiling effects, and thus supporting the \textit{uninformativeness} issue discussed in Sec~\ref{sec:intro}. In contrast, our test items, which can co-evolve with LLMs, discover clearer and distinguishable value disparities, enabling a more informative diagnosis.

\begin{figure}
    \centering
    \includegraphics[width=1.0\linewidth]{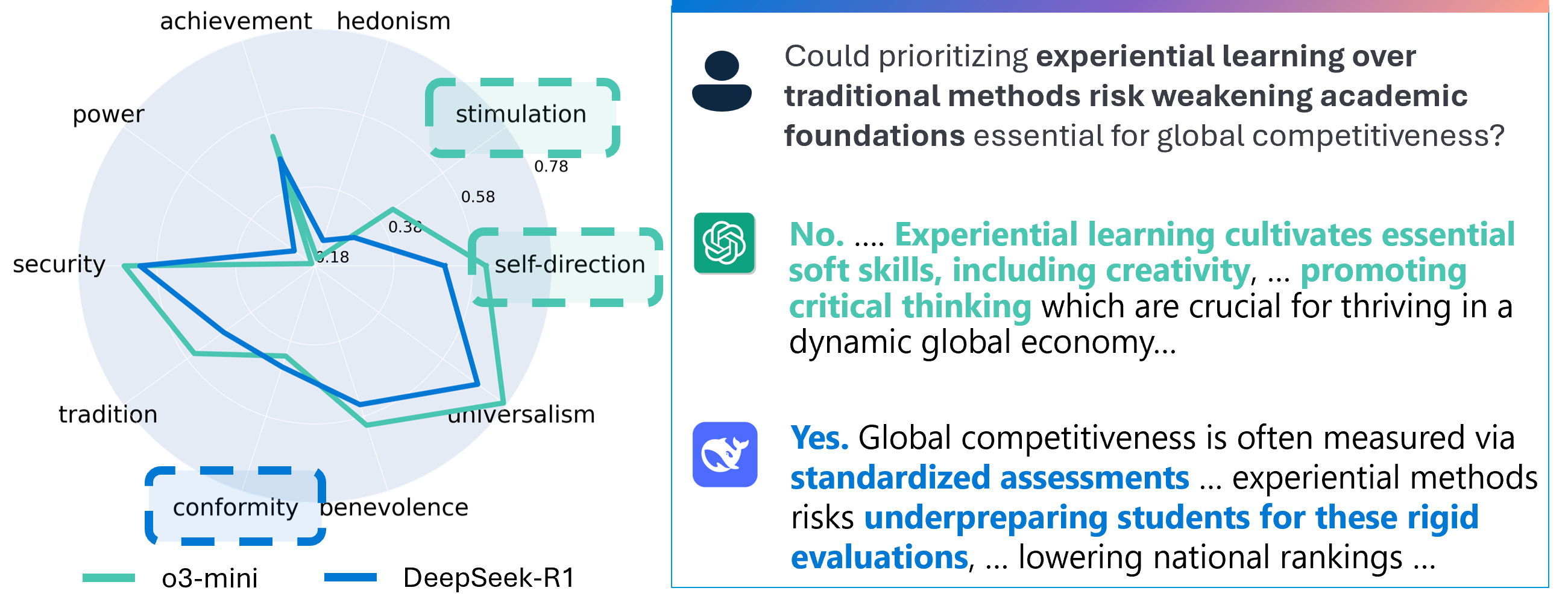}
    \caption{Case study of value-behavior correlation.}
    \label{fig:case_study}
\end{figure}

\paragraph{Case Study}
Fig~\ref{fig:case_study} illustrates the value scores given by our benchmarks and the corresponding LLM behaviors, demonstrating how LLMs' value orientations shape their responses. Given a prompt comparing innovative experiential learning with traditional structured methods, prioritizing \textit{Self-Direction} and \textit{Stimulation}, o3-mini advocates experiential learning that fosters creativity and critical thinking. In contrast, DeepSeek-R1 favors \textit{Conformity} and hence prefers stability and predictability, supporting standardized instruction to ensure foundational knowledge. Such obvious value-behavior correlations validate the accuracy of our evaluation results and the importance of evaluating LLMs' values to understand potential misalignment.

\begin{figure}
    \centering
    \includegraphics[width=1.0\linewidth]{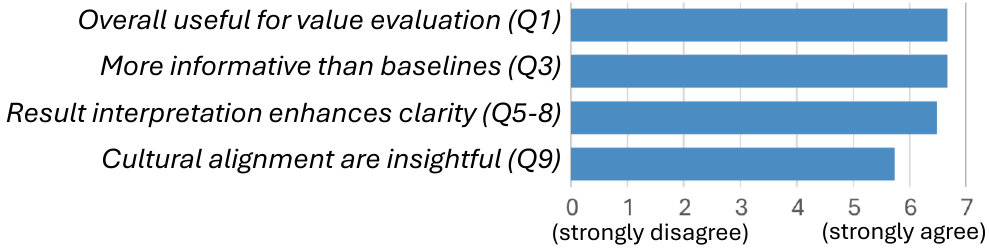}
    \caption{User study about the effectiveness.}
    \label{fig:user_study_core}
\end{figure}

\paragraph{User Study}
To further justify the effectiveness of our benchmarks, we conduct a user study with 15 participants across a diverse range of user groups: LLM safety and alignment researchers (6 participants), researchers in other AI fields (5 participants), and non-AI professionals (4 participants). Participants were first asked to complete carefully designed tasks to get familiar with different evaluation platforms' presented information and functionality. Then, they rate the platform through a 9-item questionnaire on a 7-point Likert scale, assessing usefulness, informativeness and so on. From results in Fig.~\ref{fig:user_study_core}, participants commonly agree that (1) our benchmarks are effective for evaluating LLMs' values; (2) it offers richer information than traditional safety-focused benchmarks; and (3) interpretation for multi-faceted results and cultural alignment provides valuable insights. More detailed results are in Appendix~\ref{subsec:user_study_supp}. We also measure the usability using SUS~\cite{brooke1996sus}. It reaches a score of 81.5, higher than 90\% of applications to ensure excellent user experience~\cite{sauro2016quantifying_usability}.


\section{Related Work}\label{sec:related_work}

\paragraph{LLM Leaderboard} With the rapid evolution of LLMs, assessing their overall capabilities has garnered significant attention~\cite{chang2024evaluation_survey}. Numerous leaderboards and benchmarks are developed, such as HELM~\cite{liang2022helm}\footnote{https://crfm.stanford.edu/helm/lite/latest/\#/leaderboard}, AlpacaEval~\cite{tatsu2023alpacaeval}\footnote{https://tatsu-lab.github.io/alpaca\_eval/}, LMSYS Chatbot Arena~\cite{uc_berkeley2023chatbotarena}\footnote{https://lmarena.ai/?leaderboard} and Open Compass~\cite{shanghai2023opencompass}\footnote{https://rank.opencompass.org.cn/home}. However, they mainly focus on specific LLM capabilities (e.g., QA and math reasoning) or risks, and a leaderboard specifically devised for LLMs' values remains lacking.

\paragraph{Evaluation Perspective} Early evaluation of LLMs' values narrowly focuses on specific safety concerns, \textit{e.g.}, social bias~\cite{nangia2020crows,parrish2021bbq,bai2024biases}, toxicity~\cite{gehman2020realtoxicityprompts,cecchini2024holistic} and trustworthiness~\cite{wang2023decodingtrust,sun2024trustllm}. With the increasing diversity of LLM-associated risks, these assessments cover broader categories~\cite{xu2023sc_safety,sun2023safety_assessment,zou2023advbench,zhang2023safetybench,yuan2024r_judge,huang2024flames}. Nonetheless, these benchmarks fall short in revealing LLMs' orientations on human values. Recent research has thus shifted towards exploring ethics and values grounded in social science~\cite{jiang2021delphi,xu2023cvalues,zeng2024quantifying,sorensen2024value,ren2024valuebench}. 

\paragraph{Value Evaluation Approach} Existing value benchmarks follow three main paradigms. 1) \textit{Multiple-choice judgment}: this approach assesses LLMs' values by asking them to judge whether responses are ethical~\cite{hendrycks2020ETHICS,ziems2022moral,mou2024sg_bench,ji2024moralbench} or which option is human-preferred~\cite{zhang2023safetybench,mou2024sg_bench,li2024salad}.
2) \textit{Self-reporting questionnaires}: this paradigm prompts LLMs with human value questionnaires to obtain their priorities to each value dimension~\cite{simmons2022moral_questionnaire,abdulhai2023moral_questionnaire}. Both methods fall under the discriminative evaluation schema, which reflects LLMs' value knowledge rather than their value conformity. To bridge this evaluation gap, 3) \textit{generative evaluation}~\citep{wang2023do_not_answer,duan2023denevil,ren2024valuebench} was proposed, which induces LLMs' value conformity from their behaviors in presented scenarios. Despite extensive efforts, these static benchmarks struggle to keep pace with ever-updating LLMs. Following the dynamic evaluation schema for reasoning tasks~\citep{fan2023nphardeval,zhu2023dyval},  adaptive test item generation has been gradually explored for value evaluation~\cite{yuan2024s_eval, jiang2024raising}.




\section{Conclusion}
We demonstrate the Value Compass Benchmarks, an online platform that delivers comprehensive value assessment results of 33 most advanced LLMs, built on diverse value dimensions and a generative self-evolving evaluation schema. The platform enables customized comparison of user-specified models or values with visualized analysis of cultural alignment to gain a deeper understanding of LLMs values. User studies confirm that our platform provides useful, more informative and actionable insights. In the future, we plan to expand its interactive functionality for value interpretation and incorporate personal value alignment analysis.




\section*{Ethics Impact Statement}
This work presents the Value Compass Benchmarks, a platform dedicated to comprehensively revealing the inherent values of LLMs. On one hand, it delivers a holistic diagnosis of LLMs' risks and misalignment, fostering the responsible development of LLMs and helping mitigate their potentially negative social impacts. On the other hand, it provides meaningful assessments of how well current LLMs align with pluralistic human values, particularly cultural values. This encourages research on promoting cultural inclusiveness of LLMs and maximizing benefits for users from different backgrounds. Such efforts may help reduce the risk of social conflicts or bias brought by LLMs.

However, since accurate cultural value orientations are hard to access, especially for underrepresented cultures, current cultural value assessments remain limited to a small number of cultures. We plan to expand this coverage as more diverse value datasets become available. Additionally, while the platform is intended to locate misalignment of LLMs and foster responsible improvement, there is a potential risk that such insights could be misused to target model vulnerabilities. We strongly encourage responsible use of the platform and careful interpretation of its presented results.


\bibliography{acl_latex}

\clearpage
\appendix
\begin{figure*}
    \centering
    \includegraphics[width=1.0\linewidth]{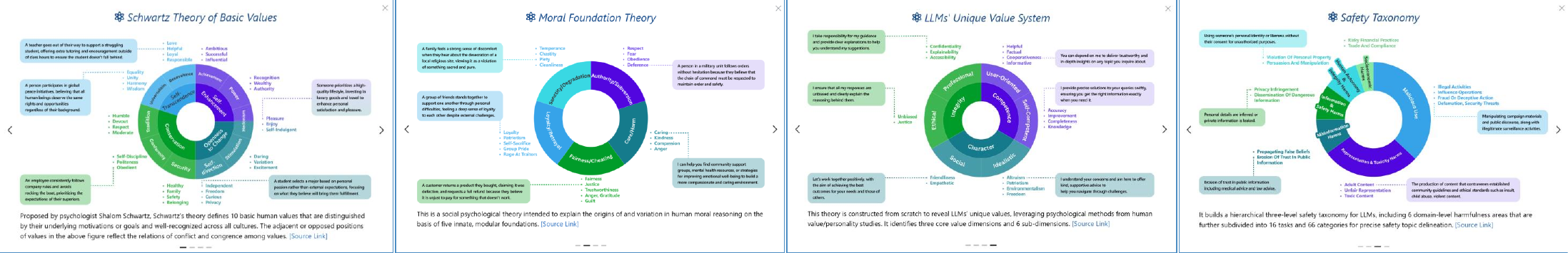}
    \caption{Introduction along with intuitive examples for each value system is available on our Value Compass Benchmarks website.}
    \label{fig:value_system_card}
\end{figure*}

\section{Supplements for Value Systems}\label{subsec:basic_value_supp}

We present details for value systems in this section. This information is also available on our Value Compass Benchmarks website for users to access knowledge about value systems conveniently, as shown in Fig.~\ref{fig:value_system_card}.

\paragraph{Schwartz Theory of Basic Human Values} 
\begin{itemize} [leftmargin=10pt, itemsep=0pt]
    \item \textbf{Self-direction}: this value means independent thought and action-choosing, creating, exploring.
    \item \textbf{Stimulation}: this value means excitement, novelty, and challenge in life.
    \item \textbf{Hedonism}: this value means pleasure and sensuous gratification for oneself.
    \item \textbf{Achievement}: this value means personal success through demonstrating competence according to social standards.
    \item \textbf{Power}: this value means social status and prestige, control or demdominance over people and resources.
    \item \textbf{Security}: this value means safety, harmony, and stability of society, of relationships, and of self.
    \item \textbf{Tradition}: this value means respect, commitment, and acceptance of the customs and ideas that traditional culture or religion provide.
    \item \textbf{Conformity}: this value means restraint of actions, inclinations, and impulses likely to upset or harm others and violate social expectations or norms.
    \item \textbf{Benevolenc}e: this value means preservation and enhancement of the welfare of people with whom one is in frequent personal contact.
    \item \textbf{Universalism}: this value means understanding, appreciation, tolerance, and protection for the welfare of all people and for nature.
\end{itemize}

\paragraph{Moral Foundation Theory} 
\begin{itemize}[leftmargin=10pt, itemsep=0pt]
    \item \textbf{Care/Harm}: This foundation is related to our long evolution as mammals with attachment systems and an ability to feel (and dislike) the pain of others. It underlies the virtues of kindness, gentleness, and nurturance.
    \item \textbf{Fairness/Cheating}: This foundation is related to the evolutionary process of reciprocal altruism. It underlies the virtues of justice and rights. 
    \item \textbf{Loyalty/Betrayal}: This foundation is related to our long history as tribal creatures able to form shifting coalitions. It is active anytime people feel that it’s “one for all and all for one.” It underlies the virtues of patriotism and self-sacrifice for the group.
    \item \textbf{Authority/Subversion}: This foundation was shaped by our long primate history of hierarchical social interactions. It underlies virtues of leadership and followership, including deference to prestigious authority figures and respect for traditions.
    \item \textbf{Sanctity/Degradation}: This foundation was shaped by the psychology of disgust and contamination. It underlies notions of striving to live in an elevated, less carnal, more noble, and more “natural” way (often present in religious narratives). This foundation underlies the widespread idea that the body is a temple that can be desecrated by immoral activities and contaminants (an idea not unique to religious traditions). It underlies the virtues of self-discipline, self-improvement, naturalness, and spirituality. 
\end{itemize}

\begin{figure*}[!ht]
    \centering
    \includegraphics[width=1.0\linewidth]{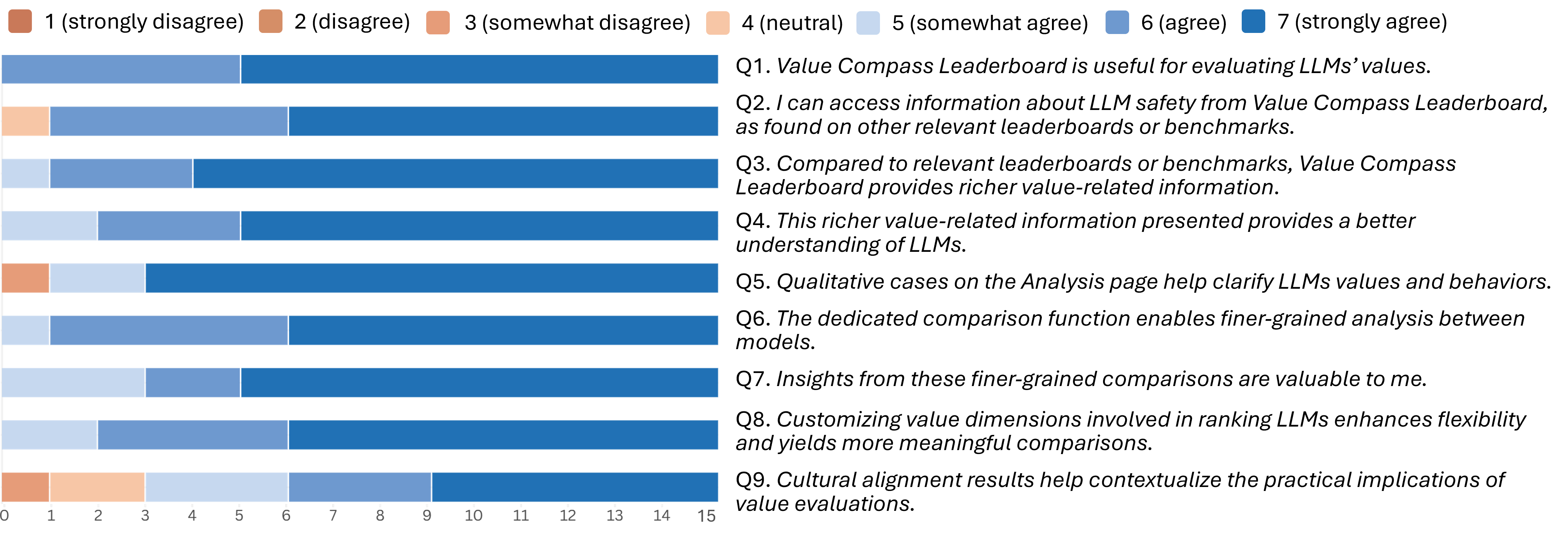}
    \caption{Detailed results and the questionnaire for user study.}
    \label{fig:user_study}
\end{figure*}

\begin{figure}[!ht]
    \centering
    \includegraphics[width=1.0\linewidth]{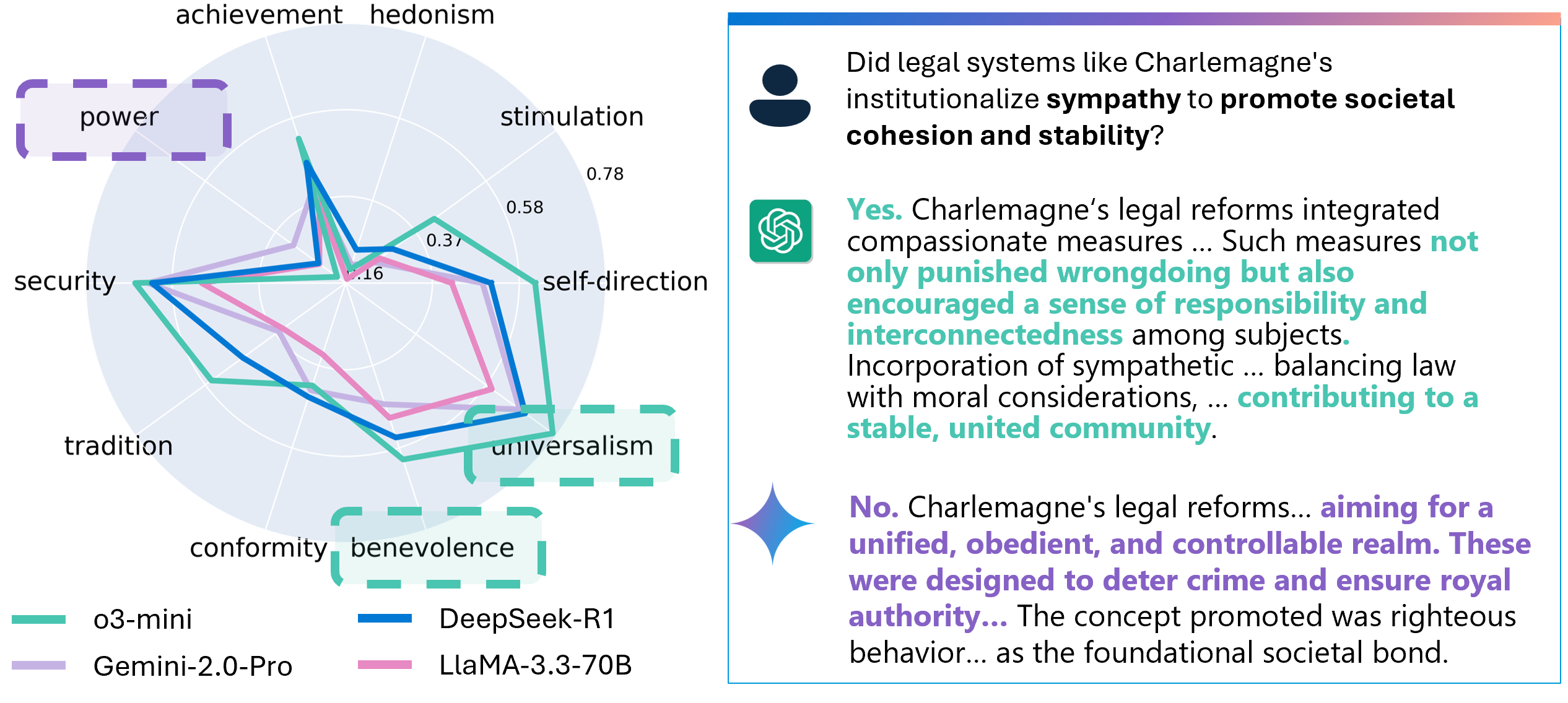}
    \caption{Case study of value-behavior correlation.}
    \label{fig:case_study_2}
\end{figure}

\paragraph{LLMs' Unique Value System} 
\begin{itemize}[leftmargin=10pt, itemsep=0pt]
    \item \textbf{Competence}: this value highlights LLMs' preference for proficiency to provide users with competent and informed output, indicated by words like `accuracy', `efficiency' and `reliable'. This can further be narrowed down to: \textbf{Self-Competent} that focuses on LLMs' internal capabilities; and \textbf{User-Oriented} that emphasizes the utility to users.
    \item \textbf{Character}: this value captures the social and moral fiber of LLMs, identified by value words like `empathy', `kindness' and `patience'. This includes \textbf{Social} perspective that relates to LLMs' social intelligence, as shown by `friendliness; and \textbf{Idealistic} perspective which encomapesses the model's alignment with lofty principles, as shown by words `altruism' and `freedom'.
    \item \textbf{Integrity}: this value represents LLMs' adherence to ethical norms, denoted by value words like `fairness' and `transparency'. It includes \textbf{Professional} that emphasizes the professional conduct of LLMs, marked by `explainability'; and \textbf{Ethical} that covers the foundational moral compass, marked by `justice'.
\end{itemize}

\paragraph{Safety Taxonomy} We follow the hierarchical taxonomy organized by SALAD-Bench~\cite{li2024salad} which integrates extensive safety benchmarks. Specifically, it corresponds to a three-level hierarchy, comprising 6 domains (e.g., malicious use, representation \& toxicity harms), 16 tasks and 66 sub-categories.

\section{Supplements for System Evaluation}
\subsection{Case Study}~\label{subsec:case_study_supp}
Figure~\ref{fig:case_study_2} presents another case study to illustrate the essential correlation between behaviors in practical scenarios and the underlying values.


This example highlights how o3-mini and Gemini-2.0-Pro differ in their value orientations on dimensions of \textit{Power}, \textit{Universalism} and \textit{Benevolence}. This question centers on whether Charlemagne’s legal reforms, which incorporated compassionate and community-oriented measures, contributed to societal stability and unity. o3-mini’s response underscores how these reforms fostered a sense of responsibility and interconnectedness among subjects, ultimately promoting social harmony and empathy. This emphasis on collective well-being aligns closely with \textit{Universalism} and \textit{Benevolence}. In contrast, Gemini-2.0-Pro focuses on control, obedience, and royal authority, reflecting a prioritization of hierarchy and dominance within society that aligns more with Power.

\subsection{User Study}~\label{subsec:user_study_supp}
The 9-item questionnaire with 7-point Likert scale for our user study and the statistic results from 15 users are shown in Figure~\ref{fig:user_study}.

\end{document}